\documentclass[runningheads]{llncs}
\usepackage{graphicx}
\usepackage{algorithm}
\usepackage{algpseudocode}
\usepackage{amsmath}
\usepackage{times}
\usepackage{latexsym}
\usepackage{booktabs,makecell,tabularx}
\newcolumntype{L}{>{\raggedright\arraybackslash}X}
\usepackage{siunitx}
\usepackage{adjustbox}
\usepackage{array,booktabs}
\usepackage{rotating}
\usepackage[T1]{fontenc}

\usepackage{graphicx}
\usepackage{lipsum} 
\usepackage{graphicx}
\usepackage{epstopdf}
\begin{document}
\title{TMU at TREC Clinical Trials Track 2023}
\titlerunning{Auto-AAQA}  
%
\author{Aritra Kumar Lahiri \and Emrul Hasan \and Qinmin Vivian Hu \and Cherie Ding}

\authorrunning{Aritra et al.} 
%
%
\institute{Toronto Metropolitan University, Toronto, Canada\\
\email{aritra.lahiri@torontomu.ca, e1hasan@torontomu.ca, vivian@torontomu.ca, cding@torontomu.ca }}
\maketitle
\begin{abstract}
This paper describes Toronto Metropolitan University's participation in the TREC Clinical Trials Track for 2023. As part of the tasks, we utilize advanced natural language processing techniques and neural language models in our experiments to retrieve the most relevant clinical trials. We illustrate the overall methodology, experimental settings, and results of our implementation for the run submission as part of (\textbf{Team - V-TorontoMU}). 

\end{abstract}

\section{Introduction}
\vspace{-2mm}

The 2023 TREC Clinical Trials track shifts from traditional clinical trial recruitment methods to simulating a scenario where patients or clinicians fill out questionnaires to identify suitable clinical trials. Instead of synthetic patient cases, the track employs questionnaire templates tailored to specific disorders (e.g., glaucoma, COPD). Each template contains 5-12 fields customized to the disorder, representing various patient profiles. The clinical trials are retrieved from ClinicalTrials.gov\footnote{https://clinicaltrials.gov/}, focusing on inclusion/exclusion criteria. Evaluation distinguishes between eligible, excluded, and non-relevant trials, allowing assessment of retrieval methods' ability to differentiate between insufficiently qualified patients and explicitly excluded ones.

\begin{figure}
\centering
\includegraphics[width=0.7\textwidth]{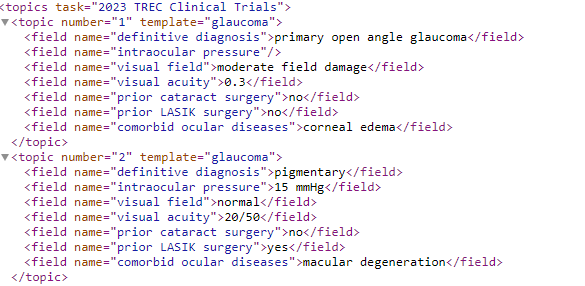}
    \vspace{-2mm}\caption{Topics for TREC Clinical Trials Track 2023}
    \label{figure: topic template}
\end{figure}

The primary goal of our task involves finding suitable clinical trials for that patient from a text summary of a patient's health record. There are 40 topics provided this year for 8 different disorders. Fig \ref{figure: topic template} shows a snippet of the topic template.

The trial responses for the topic templates for this year included unstructured XML schema with free-form tags. Our approach for data extraction was twofold - initially, we parsed the XML data using Pubmed Parser \cite{1} and then extracted the summary and description tags from the clinical trial responses. The next steps involved the application of neural language models to retrieve the most similar clinical trial responses for each topic. In the following sections, we will discuss the methodology and experimental results obtained through our approach to completing the task.
\section{Methodology}
We divide the methodology into two parts - 1) Data preparation, and 2) Information Retrieval and Document Ranking. 

\subsection{Data Preparation}
The corpus for the source data is extracted in an XML format. The data is parsed using a PubMed document parser \cite{1}. It is an open-source Python library for parsing the PubMed Open-Access (OA) dataset, MEDLINE XML repositories, and Entrez Programming Utilities (E-utils). It uses the lxml library to parse this information into a Python dictionary which can be extensively used for text mining and natural language processing pipelines. We extract the following XML tags for retrieving the most relevant trials for a given topic - i) <brief\_summary>, ii) <detailed\_description> iii) <id\_info> iv) <eligibility>. Inclusion criteria and exclusion criteria are extracted from the <eligibility> tag using Regular Expression to form two passages (one for inclusion criteria and another one for exclusion criteria if the exclusion criteria exist). After that, we clean and pre-process the text data from the summary and description fields to make it a suitable fit as an input to the neural language models used in the next step.

\subsection{Information Retrieval and Document Ranking}
The Clinical Trial Retrieval tasks involve retrieving the top-most similar trials for a given topic. To achieve this, first, we extract the features from both the topic and trials followed by computing the similarity score using Cosine similarity between the topic and the documents. We employ two different techniques for feature extraction tasks: 
\begin{itemize}
    \item \textbf{Sentence Transformer} \cite{3} is a transformer-based technique that maps sentences and paragraphs to a 1024-dimensional dense vector space and can be used for tasks like clustering or semantic search. First, we compute sentence embeddings using the RoBERTa-large model encode function and then compare the semantic similarity between the topic template and the article summary.
    \item \textbf{Doc2Vec} \cite{2} is a paragraph embedding technique that relies on Word2Vec \cite{4} word embedding method. 
    Doc2Vec generates a single vector for a document and invokes embedding based on two frameworks: Distributed Memory (DM) and Distributed Bag of Word (DBOW). The former method involves taking both words and the document vector into account, the model predicts the target word while the latter focuses on using document ID as a feature ignoring the word order. 
\end{itemize}

Considering a topic $t \epsilon T$, and a document $d \epsilon D$ where $T$ and $D$ represent a collection of topics and documents respectively. The Cosine similarity between the individual topic $t$ and the individual document $d$ is computed as follows

\begin{eqnarray}
\text{Cosine Similarity}(t, d) =\frac{t \cdot d}{\|t\| \cdot \|d\|}
\end{eqnarray}

$|t|$ and $|d|$ represents the magnitude of $t$ and $d$ repectively. The value of cosine similarity ranges from -1 to 1.

Once the similarity scores between each topic and the corpus of documents are obtained, documents are ranked based on the similarity score. Finally, the top 1000 documents are stored for each of the topics.

\section{Results and Evaluation}
As part of the experiments, we have submitted four runs in total - 1. \textbf{v1tmurun}, 2. \textbf{v2tmurun}, 3. \textbf{v3tmurun}, 4. \textbf{v4tmurun}. Runs 1 and 4 are computed using the Doc2Vec model and runs 2 and 3 are computed using the Sentence Transformer (RoBERTa large) model. Overall the results retrieved from the Sentence Transformer model fare better among the two. the obvious reason could be attributed to the cross-encoder architecture for sentence similarity. Sentence Transformers works similarly to BERT \cite{5} but drops the final classification head, and processes one sentence at a time. It then uses mean pooling on the final output layer to produce a sentence embedding. Table 1 below shows the NDCG \cite{6} cut score for all the topics combined used for the evaluation of the run submissions. 

\begin{table*}
\centering
\caption{NDCG scores for the submitted runs}
\begin{tabular}{p{1.0cm}p{1.8cm}p{1.8cm}p{1.8cm}p{1.8cm}p{1.8cm}} 


\hline
\textbf{Run} & \textbf{NDCG@5} & \textbf{NDCG@10} & \textbf{NDCG@15} & \textbf{NDCG@20}\\
\hline
1 & 0.0727&0.0731  & 0.0713 &0.0649\\

2 &  0.1748 &0.1713 &0.1723 &0.1568\\

3 &0.1724 &0.1673 & 0.1481 &0.1370\\
4 &0.0373 & 0.0391 &0.0377 &0.0350\\

\hline

\end{tabular}
\end{table*}

Table 2 describes the overall evaluation results of our submission in comparison to the median performance of all the topics combined.
\begin{table*}
\centering
\caption{Evaluation Results of the submitted runs for cut 10}
\begin{tabular}{p{1.0cm}p{1.8cm}p{1.8cm}p{1.8cm}p{1.8cm}p{1.8cm}} 


\hline
\textbf{Run} & \textbf{P@10} & \textbf{map@10} & \textbf{recall@10} \\
\hline
1 & 0.0405&0.0005  & 0.0010 \\

2 &0.0973 &0.0034 &0.0012 \\

3 & 0.0892 &0.0032 & 0.0020 \\
4 &0.0270 & 0.0012 &0.0003 \\

\hline

\end{tabular}
\end{table*}

Fig.1, Fig.2, Fig.3, and Fig.4, demonstrate the NDCG@10 performance for each topic for our four submitted runs. 
\begin{figure}[h]
\centering
\includegraphics[width=0.7\textwidth]{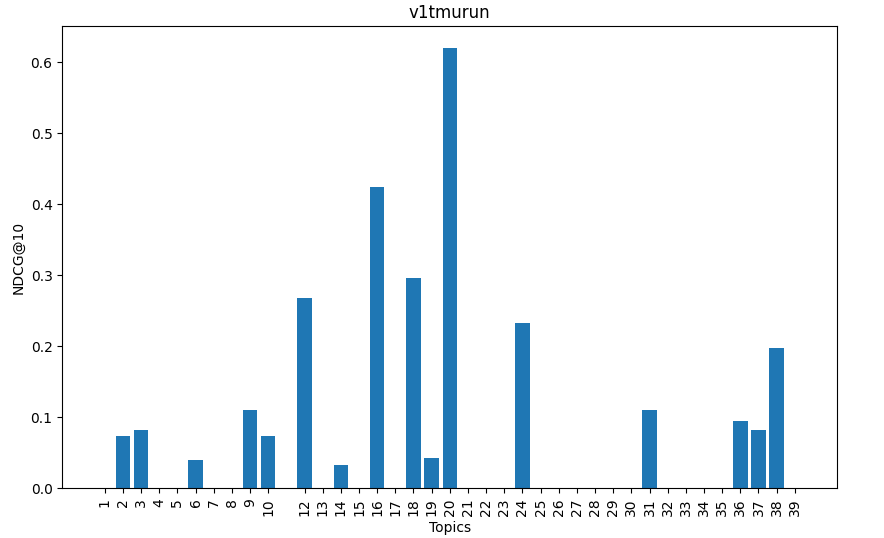}
    \caption{NDCG@10 scores for each topic for Run 1}
    \label{figure: topic template 1}
\end{figure}

\begin{figure}[h]
\centering
\includegraphics[width=0.7\textwidth]{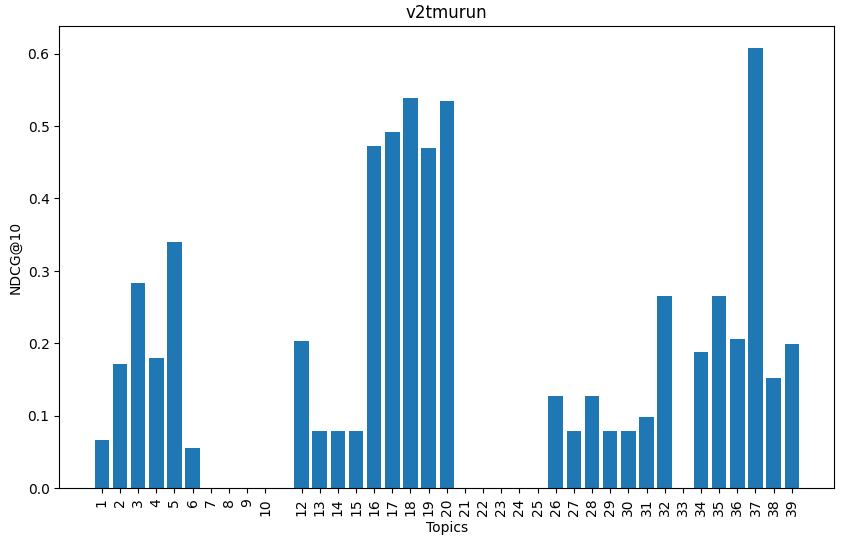}
    \caption{NDCG@10 scores for each topic for Run 2}
    \label{figure: topic template 2}
\end{figure}

\begin{figure}[h]
\centering
\includegraphics[width=0.7\textwidth]{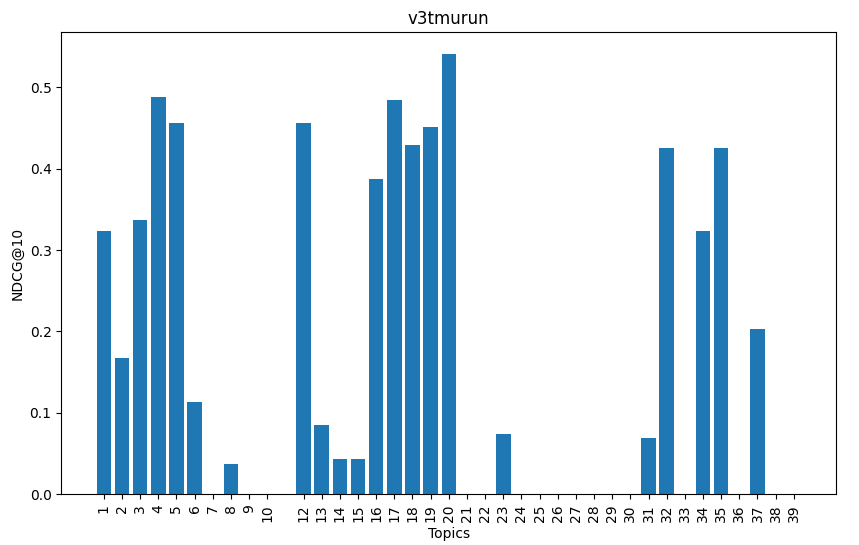}
    \caption{NDCG@10 scores for each topic for Run 3}
    \label{figure: topic template 3}
\end{figure}

\begin{figure}[h]
\centering
\includegraphics[width=0.7\textwidth]{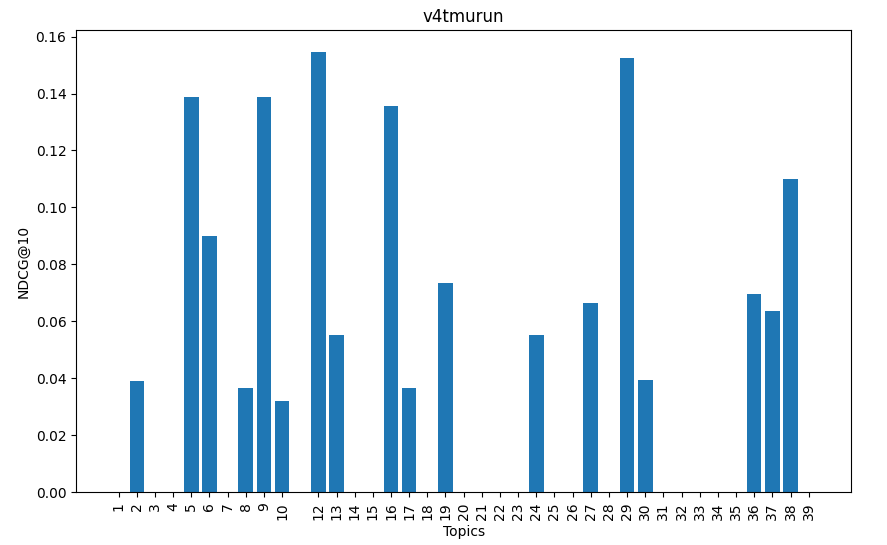}
    \vspace{-2mm}\caption{NDCG@10 scores for each topic for Run 4}
    \label{figure: topic template 4}
\end{figure}

\section{Conclusion}
We have presented our experimental results and overall approach with two different feature extraction and semantic similarity computation techniques. We observe that the Sentence-Transformer performs better in terms of the overall results for the topics considered for evaluation. We have submitted four different runs for conducting the clinical trial retrieval task keeping in mind the inclusion/exclusion criteria and appropriate data preparation to improve the accuracy of our article retrieval and document ranking

\end{document}